\RequirePackage{fix-cm}
\documentclass[smallextended]{svjour3}       
\smartqed  
\usepackage{graphicx}
\usepackage[utf8]{inputenc}
\usepackage{tikz}
\usepackage{multirow}
\usepackage{booktabs}

\begin{document}

\title{Atalaya at TASS 2019: Data Augmentation and Robust Embeddings for Sentiment Analysis
}


\author{Franco M. Luque}


\institute{F. Luque \at
              Universidad Nacional de C\'ordoba \& CONICET \\
              \email{francolq@famaf.unc.edu.ar}}

\date{Received: date / Accepted: date}

\maketitle

\begin{abstract}
In this article we describe our participation in TASS 2019, a shared task aimed at the detection of sentiment polarity of Spanish tweets.
We combined different representations such as  bag-of-words, bag-of-characters, and tweet embeddings.
In particular, we trained robust 
subword-aware word embeddings and computed tweet representations using a weighted-averaging strategy.
We also used two data augmentation techniques to deal with data scarcity: two-way translation augmentation, and instance crossover augmentation, a novel technique that generates new instances by combining halves of tweets.
In experiments, we trained linear classifiers and ensemble models, obtaining highly competitive results despite the simplicity of our approaches.
\end{abstract}

\section{Introduction}

TASS is a shared task organized every year, since 2012, with challenges related to Sentiment Analysis in Spanish.
In TASS 2019 \cite{overview_tass2019}, the proposed task is to label tweets according to the general sentiment polarity they express, classifying them into four classes: P (positive), N (negative), NEU (neutral, undecided) and NONE (no sentiment).

Five datasets are offered for the task, each one from a different Spanish speaking country: CR (Costa Rica), ES (Spain), MX (México), PE (Perú) and UY (Uruguay).
Each corpus is divided into train, development and test sections.
No other supervised datasets can be used, but external linguistic resources such as embeddings and lexicons are allowed.

The challenge is divided into two subtasks.
In monolingual subtask 1, systems must be trained and tested on the same dataset.
In cross-lingual subtask 2, systems must be trained using datasets from countries others than the one used for testing.

In this article, we describe our participation in TASS 2019 as team Atalaya.
We based our systems on our previous work \cite{atalaya_tass2018} for TASS 2018 \cite{overview_tass2018}.
For this edition, we focused our work on data augmentation and robust representations.

To represent tweets, we used a combined approach of bag-of-words, bag-of-characters and tweet embeddings.
Tweet embeddings were computed from word embeddings using a weighted averaging scheme.
For word embeddings, we used fastText subword-aware vectors \cite{bojanowski16} specifically trained for sentiment analysis over Spanish tweets.

Our fastText embeddings are robust to noise since they can compute embeddings for unseen words by using subword embeddings.
Moreover, we trained them using a database of 90M tweets from various Spanish-speaking countries, giving wide domain-specific vocabulary coverage.
We achieved additional robustness by doing preprocessing using several text normalization and noise reduction techniques.

To cope with training data scarcity, we experimented with data augmentation techniques.
As in our previous work, we did augmentation using machine translation to and from several other languages.

We also tried a novel augmentation technique we called instance crossover, loosely inspired by the crossover operation from genetic algorithms.
This technique combines halves of tweets to generate new instances.
Despite its simplicity, this idea showed to be useful in our experiments.

For the classifying models, we used logistic regressions and also bagging ensembles of logistic regressions.


The rest of the paper is as follows.
The next section presents the components of our systems and the ideas we used to build them.
Section 3 presents the experiments and results for both subtasks.
Section 4 concludes the work with some observations about our experience.

\section{Techniques and Resources}
\label{sec:techniques}

\begin{figure}[tp]
\tiny
\centering 
\tikzstyle{sensor}=[draw, fill=blue!20, text width=5em, 
    text centered, minimum height=2.5em]
\tikzstyle{naveqs} = [sensor, text width=6em, fill=red!20, 
    minimum height=12em, rounded corners]
\begin{tikzpicture}
\node[sensor] (train) {Training Data};
\node[sensor,below of=train] (aug1) {Translation Augmented Data};
\node[sensor,below of=aug1] (aug2) {Crossover Augmented Data};
\node[sensor] (pre) at (2.25,-1.5) {Preprocess};
\node[sensor,right of=train,node distance=4.5cm] (boc) {BoC};
\node[sensor,below of=boc] (bow) {BoW};
\node[sensor,below of=bow] (emb) {Embedding}; 
\node[sensor,right of=bow,node distance=2cm] (clf) {Classifier}; 

\draw[->,dotted] (train) -- ++(-1,0)-- ++(0,-1) -- (aug1);
\draw[->,dotted] (train) -- ++(-1,0)-- ++(0,-2) -- (aug2);

\draw (train) -- ++(1,0) -- ++(0,-2) -- (aug2);
\draw[->] (aug1) -- ++(1.25,0) -- ++(0,-0.5) -- (pre);
\draw[->] (aug1) -- ++(1.25,0) -- ++(0,1) -- (boc);

\draw[->] (pre) -- ++(1.25,0) -- ++(0,0.5) -- (bow);
\draw[->] (pre) -- ++(1.25,0) -- ++(0,-0.5) -- (emb);

\draw (boc) -- ++(1,0) -- ++(0,-2) -- (emb);
\draw[->] (bow) -- (clf);

\draw[->] (clf) -- ++(1,0);
\end{tikzpicture}
 \caption{
General architecture of our systems, including the input used in training time.
Dotted lines denote precomputed processes.
}
 \label{fig:system}
\end{figure}
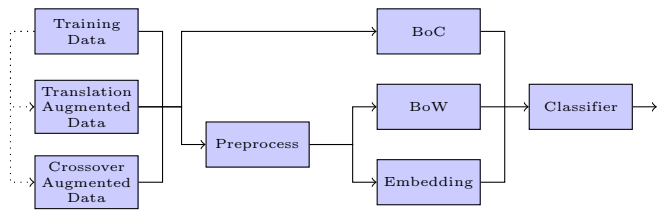

The components and general architecture of our systems is shown in Fig.~\ref{fig:system}.
In this section, we describe the techniques and resources we used to build them.

\subsection{Preprocessing}
\label{sec:preprocessing}

Preprocessing is important to reduce noise from tweets.
We follow our previous work, applying two levels of preprocessing.
Basic tweet preprocessing includes tokenization, replacement of handles, URLs, and e-mails, and shortening of repeated letters.
Further preprocessing is done, aimed at semantic tasks.
It includes removal of punctuation, stopword
and numbers, lowercasing, lemmatization, and negation handling.

For negation handling, we followed a simple approach \cite{das01,pang02}: We
find negation words and add the prefix \texttt{NOT\_}
to the following tokens. Up to three tokens are
negated, or less if a non-word token is found.

No treatment was performed to hashtags, emojis, interjections and onomatopoeias.
Moreover, no spelling correction nor any other additional normalization was applied.

\subsection{Bags of Words and Characters}

A simple way to represent textual data as feature vectors is to use bag-of-words (BoWs).
A bag-of-words represents a tweet as a vector with the counts of words occurring in it.
Resulting vectors are high-dimensional and sparse.
The BoW representation can be extended to count also word n-grams.
In this work, we used BoWs, together with count binarization and TF-IDF re-weighting, both useful for semantic tasks such as sentiment analysis.


For more robustness, we also used a bag-of-characters
(BoC) representation.
BoCs have exactly the same properties and variants than BoWs but are applied to characters instead of word tokens.
Character usage in tweets holds useful information for sentiment analysis.
In our work, the BoC representation is computed over the original raw text of tweets, with no preprocessing at all.

\subsection{Word Embeddings}

A more interesting way to represent text is using embeddings.
Word embeddings are low-dimensional dense vector representations of words.
These representations are learned in an unsupervised fashion using large quantities of plain text, providing high vocabulary coverage.

For our systems, we used fastText embeddings \cite{bojanowski16},
that introduces additional robustness by learning also subword-level embeddings and using them to compute vectors for unseen words.
With subword-aware embeddings, the need for normalization of highly noisy text in preprocessing is greatly alleviated.

We did not use a pretrained fastText model but trained our own using a big preprocessed dataset of $\sim90$ million tweets from various Spanish-speaking countries.
This dataset is mostly composed of tweets we collected for previous work, and also includes the tweets from all sections of all TASS 2019 datasets.


\subsection{Tweet Embeddings}
\label{sec:sif}

To use word embeddings in sentiment analysis, the embeddings of the individual tokens must be aggregated in 
some way to obtain a complete tweet representation.

A simple approach is to do averaging to obtain a single vector.
A bit more interesting is to add weights to the averaging scheme.
This way, some words may be considered more relevant than others for the classification task.

In this work, we used Smooth Inverse Frequency (SIF), a simple weighted averaging scheme from \cite{arora17} inspired by TF-IDF re-weighting.
In SIF, words $w$ are weighted with $\frac{a}{a + p(w)}$, where $p(w)$ is the word unigram probability, and $a > 0$ is a smoothing hyperparameter.
Big values of $a$ mean more smoothing towards plain averaging.
We model the unigram probability using unigram counts from our preprocessed $\sim90$ million tweets dataset.

In \cite{arora17} a final transformation is applied to tweet embeddings by subtracting from them a common component shared by all the vectors.
Preliminary experiments with this idea, however, showed it to be harmful to our systems, so we did not use it in our final experiments.

An important limitation of this tweet embedding scheme is that word order is completely ignored.
Only preprocessing may allow the influence of ordering in the result.
In particular, the negation handling trick from section~\ref{sec:preprocessing} is a useful, although naive, way to let words be affected by previous negations.

\subsection{Data Augmentation with Two-Way Translation}
\label{sec:augmentation}

One of the main successful approaches from our previous work on TASS was the use of data augmentation techniques.
Data augmentation helps to cope with training data scarcity.
Augmentation aims at the introduction of data variability using label-preserving transformations on real data.
When correctly used, it contributes to data robustness and acts as a regularizer for the models.

Our approach for TASS 2018 (also as team Atalaya) was to use two-way translation augmentation.
In two-way translation, an external machine translation service is used to translate tweets to other ``pivot'' languages and then back to Spanish.
This augmentation technique helps to introduce lexical and syntactical variations to tweets, most times preserving their meaning.

In \cite{atalaya_tass2018} we used two-way translation to augment the training data using four pivot languages
(English, French, Portuguese and Arabic).
This augmentation was found to be useful for the ES and CR datasets, but not for PE.

In this work, we explored two-way translation further, applying it to all the datasets using 20 different pivot languages.
To get translations, we used Google's Cloud Translation API service.

Pivot languages were selected by hand from the list of available languages that the API can translate from/to Spanish.
The selection was done trying to pick representative languages from different language families.




\subsection{Data Augmentation with Instance Crossover}
\label{sec:naive_augmentation}


We also tried a new augmentation idea that aims at the generation of new data by combining pairs of instances with the same label.
We call it the instance crossover augmentation technique, inspired by the chromosome crossover operation from genetic algorithms.

Our approach is simply to split tokenized tweets into two halves, and then randomly sample and combine first halves with second halves.
Resulting instances will probably be ungrammatical and semantically unsound, but our hypothesis is that what is left of semantics, for instance at the lexical level, will preserve sentiment polarity most of the times.\footnote{
Grammaticality and semantic soundness are already rare in the original tweets, so it is not something we should worry about very much.
}

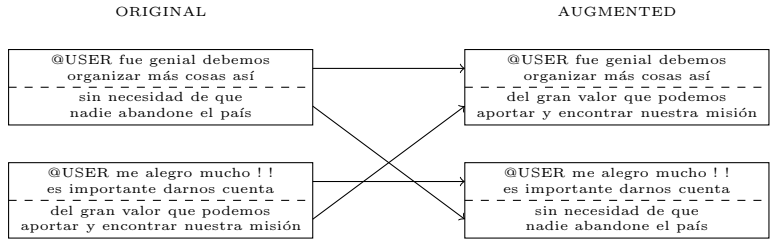
\begin{figure}[tp]
\tiny
\centering 
\begin{tikzpicture}

\node at (2,1.5) {ORIGINAL};
\node at (8,1.5) {AUGMENTED};

\node[align=center] (a) at (2,0.75)
{@USER fue genial debemos\\ organizar más cosas así};
\node[align=center] (b) [below of=a,node distance=0.5cm]
{sin necesidad de que\\ nadie abandone el país};
\draw (0,0) rectangle +(4,1);
\draw[dashed] (0,0.5) -- +(4,0);

\node[align=center] (c) at (2,-0.75)
{@USER me alegro mucho ! !\\ es importante darnos cuenta};
\node[align=center] (d) [below of=c,node distance=0.5cm]
{del gran valor que podemos\\ aportar y encontrar nuestra misión};
\draw (0,-1.5) rectangle +(4,1);
\draw[dashed] (0,-1) -- +(4,0);

\node[align=center] (a2) at (8,0.75)
{@USER fue genial debemos\\ organizar más cosas así};
\node[align=center] (b2) [below of=a2,node distance=0.5cm]
{del gran valor que podemos\\ aportar y encontrar nuestra misión};
\draw (6,0) rectangle +(4,1);
\draw[dashed] (6,0.5) -- +(4,0);

\node[align=center] (c2) at (8,-0.75)
{@USER me alegro mucho ! !\\ es importante darnos cuenta};
\node[align=center] (d2) [below of=c2,node distance=0.5cm]
{sin necesidad de que\\ nadie abandone el país};
\draw (6,-1.5) rectangle +(4,1);
\draw[dashed] (6,-1) -- +(4,0);

\draw[->] (4,0.75) -- (6,0.75);
\draw[->] (4,0.25) -- (6,-1.25);
\draw[->] (4,-0.75) -- (6,-0.75);
\draw[->] (4,-1.25) -- (6,0.25);
\end{tikzpicture}

 \caption{
Instance crossover augmentation example using two tweets with positive (P) sentiment polarity.
The original tweets are on the left.
The first half of one tweet is combined with the second half of the other, resulting in the new instances on the right.
The dotted lines show the division in halves.
}
 \label{fig:crossover}
\end{figure}

Fig.~\ref{fig:crossover} shows an example of instance crossover using two tweets with positive sentiment.
In this example, crossover is successful in the sense that the resulting instances can be clearly judged as having a positive sentiment.
In other cases, crossover may fail to preserve polarity, for instance, because of an unfortunate combination involving a negation.
Resulting instances may even be completely nonsensical, introducing noise to the data.

For this work, we chose to directly validate in experiments this augmentation idea.
In our experiments, we applied augmentation over the training tweets after basic preprocessing and before semantic preprocessing (as defined in section~\ref{sec:preprocessing}).
We tried different levels of augmentation, multiplying the size of original training datasets by factors of 4, 8, 12, 16 and 20.
We preserved the original distribution over labels and therefore the class imbalance.

Instance crossover is a very rough and naive augmentation technique.
However, it may be useful to introduce more data variability than two-way translation.
With translation, new data points may fall very close to the original ones, while crossover introduces new points in the ``spaces'' between the original ones.
Moreover, this is done in a representation agnostic fashion.
It can be used with bag-of-words, embeddings, or even neural based representations.

Another clear advantage of instance crossover is that it does not rely on any external resource or system.
Unlike this, translation requires an external service, at a cost, and other techniques such as synonym replacement require thesauruses or word similarity databases.



\section{Experiments}
\label{sec:experiments}

In this section, we describe our experiments.
We implemented all our systems using scikit-learn \cite{scikit-learn}.
In the preprocessing stage, we used an NLTK-based tokenizer \cite{bird2004nltk} and TreeTagger for lemmatization \cite{schmid95}.

\subsection{System Development}
\label{sec:development}

For simplicity, most of our work was centered on subtask 1 and on the ES dataset, looking for model configurations and hyperparameter values that gave the best results over the development section of the ES dataset.
The optimization process was done using a mixed approach of grid search and by-hand tuning.

We targeted the maximization of both macro-F1 and accuracy scores.
The macro-F1 score is the main metric of TASS 2019, but it is very unstable and sensible to small changes in predictions for minority classes.
On the other hand, accuracy is more stable and reliable for the development process.

As a starting point, we used our optimal model and configuration from TASS 2018 \cite{atalaya_tass2018}.
This model is a logistic regression over a combination of bag-of-words (BoW), bag-of-characters (BoC) and tweet embeddings as follows:
\begin{itemize}
    \item Augmentation: Two-way translation with English, French, Portuguese and Arabic ({\small EN+FR+PT+AR}) as pivot languages.
    \item BoW: All word n-grams for $n \leq 5$.
    \item BoC: All character n-grams for $n \leq 6$.
    \item Tweet embeddings: 50 dimension fastText vectors. Weighted averaging with $a = 0.1$.
    \item Logistic regression: liblinear solver with primal formulation, L2 regularization with inverse strength $C = 1.0$, and class-balanced reweighting.
\end{itemize}
Table~\ref{tab:baseline} shows a detailed evaluation of this baseline model.

The first idea we explored was augmentation using two-way translation with the new 20 pivot languages.
We tried adding all new data, but also adding some subsets of it, by grouping pivot languages in packs of four datasets.
However, we could not find any improvement from the original {\small EN+FR+PT+AR} augmentation, sometimes finding important drops in model quality.

Next, we explored augmentation using instance crossover.
We tried 4x, 8x, 12x, 16x and 20x factor augmentations from the original ES training corpus, with and without additional translation-based augmentation.
In every case results were improved w.r.t. not using crossover augmentation.
The best result was found for 8x augmentation.

Last, we tuned the hyperparameters of the logistic regression.
The best configuration found was a liblinear solver with primal formulation, L2 regularization with inverse strength $C = 0.2$, with no class-balanced reweighting.

We also tried an ensemble of logistic regressions by using bagging.
Bagging was found to be useful for the ES dataset.
The best configuration found was using a bag of 40 logistic regressions.

Table~\ref{tab:detail} shows a detailed evaluation for our best model for ES found following the development process.

\subsection{Subtask 1: Monolingual Experiments}

To build a submission for subtask 1, we first ran the final test on the ES dataset using the best model described in the previous section.

To build submissions for the other datasets, we followed a similar development approach, but with most hyperparameters fixed with the optimal values for ES.
We focused the optimization process in the usage of translation and crossover augmentations, in the logistic regression hyperparameters and in the usage of bagging.
Tuning was done mostly by hand and sometimes using grid-search.
Table~\ref{tab:config} shows the optimal configurations found for each dataset.

The final results for the complete submission for subtask 1 are shown in Table~\ref{tab:subtask1}.


\begin{table}[tp]
 \centering 
 \scriptsize
\begin{tabular}{lccc}
& Prec. & Rec. &  F1 \\
\hline
P    & 60.64 & 67.86 & 64.04  \\
N    & 67.82 & 80.83 & 73.76  \\
NEU  & 21.05 & 4.82 & 7.84  \\
NONE & 38.60 & 34.38 & 36.36  \\
\hline
macro avg & 47.03 & 46.97 & 47.00 \\
\hline
\multicolumn{2}{l}{Accuracy: 61.10} \\
\end{tabular}
\quad
\quad
\begin{tabular}{lrrrr}
 & P & N & NEU & NONE \\
\hline
P & 114 & 36 & 5 & 13 \\
N & 28 & 215 & 8 & 15 \\
NEU & 29 & 43 & 4 & 7 \\
NONE & 17 & 23 & 2 & 22 \\
\end{tabular}
 \caption{Subtask 1: Detailed results for our baseline system on the ES development set.
 Left: Classification report. Right: Confusion matrix.
 }
 \label{tab:baseline}
\end{table}

\begin{table}[tp]
 \centering 
 \scriptsize
\begin{tabular}{lccc}
& Prec. & Rec. & F1 \\
\hline
N    & 66.38 & 87.59 & 75.53 \\
NEU  & 50.00 & 2.41 & 4.60 \\
NONE & 60.71 & 26.56 & 36.96 \\
P    & 61.62 & 72.62 & 66.67 \\
\hline
macro avg & 59.68 & 47.30 & 52.77 \\
\hline
\multicolumn{2}{l}{Accuracy: 64.37} \\
\end{tabular}
\quad
\quad
\begin{tabular}{lrrrr}
 & P & N & NEU & NONE \\
\hline
P & 122 & 42 & 1 & 3 \\
N & 27 & 233 & 1 & 5 \\
NEU & 29 & 49 & 2 & 3 \\
NONE & 20 & 27 & 0 & 17 \\
\end{tabular}
 \caption{Subtask 1: Detailed results for our best system on the ES development set.
 Left: Classification report. Right: Confusion matrix.}
 \label{tab:detail}
\end{table}

\begin{table}[tp]
 \centering 
 \begin{tabular}{lccccc}
  & \multicolumn{2}{c}{augmentation} & \multicolumn{2}{c}{logistic regression} \\
  & translation & crossover & C & class-weight & bagging \\
\cmidrule(r){2-3} \cmidrule(r){4-5} \cmidrule{6-6}
ES & {\tiny EN+FR+PT+AR} & 8x  & 0.2 & no & 40 \\
PE & {\tiny EN+FR+PT+AR} & 4x  & 0.22 & balanced & no \\
CR & no & 8x  & 1.15 & balanced & no \\
UY & no & 8x  & 0.6 & no & no \\
MX & {\tiny EN+FR+PT+AR} & 16x & 0.125 & balanced & no \\
\end{tabular}
 \caption{Subtask 1:
 Best system configurations found for each dataset.
 Optimization was done using the development sections.
 Hyperparameters and their values are described in section~\ref{sec:development}.
 }
 \label{tab:config}
\end{table}

\begin{table}[tp]
 \centering 
 \begin{tabular}{lccccc}
  & \multicolumn{2}{c}{dev} & \multicolumn{2}{c}{test}\\
  & Acc. & M-F1 & Acc. & M-F1 & Rank \\
\cmidrule(r){2-3} \cmidrule(r){4-5} \cmidrule{6-6}
ES & 64.37 & 52.77 & 60.67 & 48.42 & 2 \\
PE & 51.41 & 47.90 & 45.36 & 45.38 & 1 \\
CR & 61.28 & 53.36 & 57.20 & 46.91 & 3 \\
UY & 61.93 & 54.81 & 60.64 & 49.86 & 3 \\
MX & 67.65 & 53.88 & 68.87 & 48.46 & 4 \\
\end{tabular}
 \caption{
Subtask 1: Final results for each dataset, on development and test sections.
Rank is the official final rank in the competition.}
 \label{tab:subtask1}
\end{table}


\subsection{Subtask 2: Crosslingual Experiments}

To build submissions for subtask 2 we did a minimal set of experiments.
For each language, we started from the optimal model configuration found for subtask 1 and then trained it using the union of the training datasets of every other language.
We then proceeded to optimize the main hyperparameters of the logistic regression, mostly doing by-hand tuning.

We did some preliminary experiments with data augmentation and bagging for the ES dataset.
However, results were not improved, so we didn't do further experimentation with the other datasets.

The final results for the complete submission for subtask 2 are shown in Table~\ref{tab:subtask2}.
Results are surprisingly good, considering that we did limited experimentation because of lack of time.

\begin{table}[tp]
 \centering 
 \begin{tabular}{lccccc}
  & \multicolumn{2}{c}{dev} & \multicolumn{2}{c}{test}\\
  & Acc. & M-F1 & Acc. & M-F1 & Rank \\
\cmidrule(r){2-3} \cmidrule(r){4-5} \cmidrule{6-6}
ES & 63.68 & 53.57 & 61.55 & 45.42 & 3 \\
PE & 41.97 & 43.71 & 54.64 & 47.42 & 1 \\
CR & 57.69 & 48.77 & 57.12 & 47.41 & 2 \\
UY & 60.91 & 53.15 & 61.20 & 51.35 & 1 \\
MX & 64.90 & 50.18 & 68.13 & 47.25 & 1 \\
\end{tabular}
 \caption{Subtask 2: Final results for each dataset, on development and test sections.
 Rank is the official final rank in the competition.}
 \label{tab:subtask2}
\end{table}

\subsection{Ablation Tests}

As a complementary post-competition experiment, we performed ablation tests for each of the components of our systems, to assess the relevance of each of the techniques used in this work.

The ablation tests were done using the best system for subtask 1 on the ES dataset.
The results are displayed in Table~\ref{tab:ablation}.

It can be seen that all the techniques have a positive impact.
Among representations, tweet embedding is the most important representation, way above BoW and BoC representations.
Also, it is interesting to observe that crossover augmentation has an impact on the F1 but not on the accuracy, indicating that it is helping mostly on the minority classes NEU and NONE.

\begin{table}[tp]
 \centering 
 \begin{tabular}{llcc}
 & & Acc. & M-F1 \\
\hline
full system & & 64.37 & 52.77 \\
\hline
\multirow{2}{*}{augmentation:} & no translation & 62.99 & 48.05 \\
& no crossover & 64.37 & 47.57 \\
\hline
\multirow{4}{*}{representation:} & no BoW & 62.99 & 49.89 \\
& no BoC & 62.65 & 50.80 \\
& no BoW+BoC & 62.13 & 48.26 \\
& no embeddings & 58.52 & 41.83 \\
\hline
classifier: & no bagging & 64.03 & 51.94 \\
\end{tabular}
 \caption{Ablation tests for several techniques used in our final system. Results are for subtask 1 on the ES development set.}
 \label{tab:ablation}
\end{table}


\section{Conclusions}
\label{sec:conclusion}

Robust representations and data augmentation play a strong role in sentiment analysis with small-sized training datasets.
As in our previous experience with TASS 2018, we are still able to obtain top ranking results without having to resort to complex models such as deep neural networks.

We observe that, for this edition of TASS, most of our work was on the application of general ML techniques, and not on particular task/domain specific engineering.
In particular, we successfully tried instance crossover augmentation, a novel technique that, despite its simplicity, showed a positive impact on results.
This idea can be useful to augment small datasets 
for other short text classification tasks without the need for external resources.


\begin{acknowledgements}
This work was partially supported by a research grant from SeCyT, Universidad Nacional de Córdoba.
\end{acknowledgements}

\bibliographystyle{iberlef/spmpsci}      
\bibliography{references.bib}   

\end{document}